\documentclass{article}

\usepackage{arxiv}

\usepackage[utf8]{inputenc}
\usepackage[T1]{fontenc}
\usepackage{hyperref}
\usepackage{url}
\usepackage{booktabs}
\usepackage{amsfonts}
\usepackage{nicefrac}
\usepackage{microtype}
\usepackage{graphicx}
\usepackage{natbib}
\usepackage{doi}

\usepackage{amsmath,amssymb,amsthm,mathtools}
\usepackage{subcaption}
\usepackage{multirow}
\usepackage{xcolor}
\usepackage[capitalize,noabbrev]{cleveref}

\newcommand{\Coh}{\mathcal{C}} 

\newcommand{\gsm}{\textsc{GSM8K}}

\title{Entropy Trajectory Shape Predicts LLM Reasoning Reliability:\\
A Diagnostic Study of Uncertainty Dynamics in Chain-of-Thought}

\author{
  Xinghao Zhao \\
  Huazhong University of Science and Technology \\
  \texttt{xinghaozhao60@gmail.com}
}

\renewcommand{\shorttitle}{Entropy Trajectory Shape Predicts LLM Reasoning Reliability}

\hypersetup{
pdftitle={Entropy Trajectory Shape Predicts LLM Reasoning Reliability},
pdfauthor={Xinghao Zhao},
pdfkeywords={LLM reliability, calibration, chain-of-thought, entropy trajectory},
}

\begin{document}
\maketitle

\begin{abstract}
Chain-of-thought (CoT) reasoning improves LLM accuracy on complex tasks,
yet reliable methods for detecting reasoning failures without expensive multi-sample
approaches remain elusive. We study whether the \emph{shape} of uncertainty dynamics
across reasoning steps---captured cheaply by sampling a handful of answer completions
at each step---predicts whether the final answer is correct.

We introduce the concept of \emph{entropy-trajectory monotonicity}: a chain is
monotone if its per-step answer-distribution entropy decreases at every step,
reflecting consistent uncertainty reduction. On \gsm{} ($n{=}300$) with
Qwen2.5-7B-Instruct, monotone chains achieve $68.8\%$ accuracy versus $46.8\%$
for non-monotone chains---a gap of $+21.9$ percentage points (Fisher's exact
$p{=}0.0005$; OR$=2.50$). Critically, the \emph{scalar} total entropy reduction
is not predictive ($\rho{=}{-}0.06$, $p{=}0.31$), revealing a
\emph{shape-over-magnitude dissociation}: it is \emph{whether} entropy decreases
at every step, not \emph{how much} it drops, that predicts correctness.
The dissociation extends to the graded signal: in full-scale runs, increasing
violation count consistently reduces accuracy on both GSM8K and MATH-500.

Beyond the 300-problem pilot, the effect persists at larger scale. On full
GSM8K ($n{=}1319$), monotone chains reach $93.2\%$ accuracy versus $81.7\%$ for
non-monotone chains ($+11.5$ pp). On MATH-500 ($n{=}500$), monotone chains reach
$63.7\%$ versus $30.4\%$ ($+33.3$ pp). Across both datasets, violation count is
negatively correlated with correctness (Spearman $\rho=-0.198$ on GSM8K and
$\rho=-0.381$ on MATH-500).

We further show that token log-probability confidence worsens in calibration
with step depth (ECE: $0.186 \to 0.312$ from step 0 to step 7), and that
entropy-trajectory monotonicity achieves $+5.8$ pp at $73.7\%$ coverage,
outperforming all scalar baselines including final-step entropy ($+2.2$ pp) and
scalar coherence ($-0.6$ pp, worse than random) at $\approx\!1{,}500$ tokens/question---
one-eighth the cost of 40-chain self-consistency.
At matched answered-set coverage, SC@3/SC@5 can be slightly higher than
monotonicity ranking, so our claim is not dominance over SC voting, but a
cheap and interpretable single-chain triage signal with favorable
accuracy-cost trade-offs.
The initial pilot findings further replicate on a second model
(Mistral-7B-Instruct-v0.3, $n{=}300$), where
monotone chains reach $72.3\%$ vs.\ $37.6\%$ for non-monotone chains
($+34.7$ pp; OR$=4.33$). Structural properties of uncertainty trajectories are
thus more informative than aggregate magnitude measures across model families.
Current evidence is strongest on numeric/discrete-answer tasks; extending to
open-domain free-form QA requires stronger semantic canonicalization.

\end{abstract}

\section{Introduction}
\label{sec:intro}

Large language models (LLMs) produce correct and incorrect answers with equal
surface fluency. In chain-of-thought (CoT) reasoning
\citep{wei2022chain,kojima2022large}, a model generates a multi-step solution
before giving a final answer---but a confident-looking chain of steps offers no
guarantee of correctness. Detecting failures cheaply, without generating many
independent samples, is an open practical problem.

\begin{figure}[t]
  \centering
  \begin{subfigure}[b]{0.48\textwidth}
    \includegraphics[width=\textwidth]{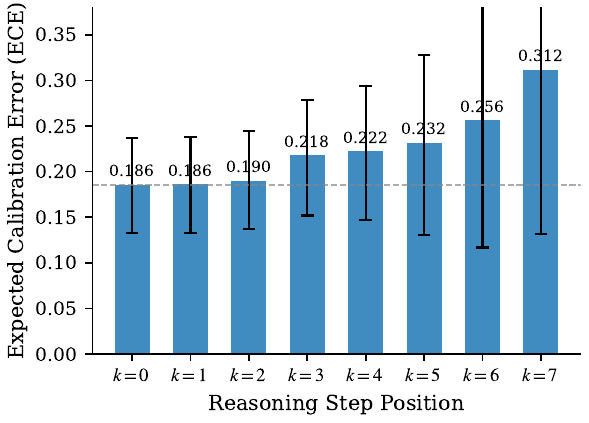}
    \caption{ECE increases monotonically with step depth (0.186 at step 0 to
    0.312 at step 7). Error bars show 95\% bootstrap CIs.}
    \label{fig:ece_by_step}
  \end{subfigure}
  \hfill
  \begin{subfigure}[b]{0.48\textwidth}
    \includegraphics[width=\textwidth]{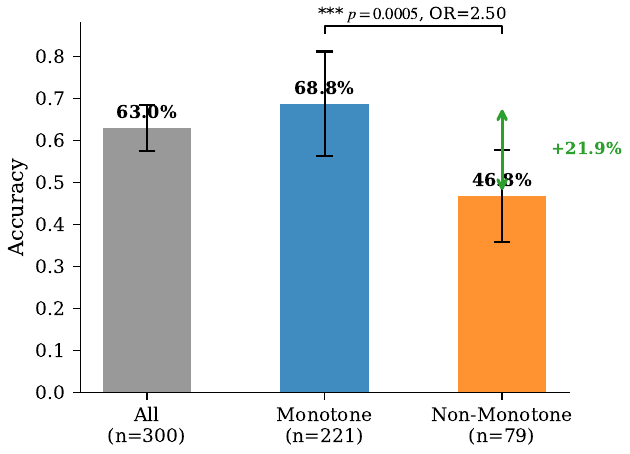}
    \caption{Monotone chains are significantly more accurate than non-monotone
    chains (OR$=2.50$, Fisher's $p{=}0.0005$, $+21.9$ pp gap).}
    \label{fig:monotone_accuracy}
  \end{subfigure}
  \caption{Two key findings from our diagnostic study.
  \textbf{(a)} Token log-probability confidence becomes increasingly
  miscalibrated at later reasoning steps.
  \textbf{(b)} Chains with monotone entropy trajectories achieve substantially
  higher accuracy than non-monotone chains.}
  \label{fig:hero}
\end{figure}

Two families of reliability signals have been studied. \emph{Self-consistency}
methods~\citep{wang2023self} aggregate answers from multiple independently sampled
chains, but require 10--40 samples per question and discard the rich step-level
information within each chain. \emph{Token log-probability} scores are cheap but
systematically miscalibrated: we show that the ECE of token log-prob confidence
increases from $0.186$ at the first reasoning step to $0.312$ at the eighth
step (\Cref{fig:hero}a), with all standard proxy functions collapsing to
near-degenerate confidence ranges. Because step-level correctness labels are
unavailable in this black-box setup, this ECE trend is interpreted as
predictive calibration for final correctness rather than step-local calibration.

We pursue a different approach: rather than scoring a chain by a single scalar
derived from token probabilities, we ask how the model's uncertainty over the
\emph{final answer} evolves step by step. At each step prefix, we sample $m{=}5$
answer completions and compute the Shannon entropy $H_k$ of the resulting answer
distribution. This gives an \emph{entropy trajectory} $(H_0, H_1, \ldots, H_N)$
at low additional cost.

The central question we study is: does the \emph{shape} of this trajectory
predict correctness, beyond what its \emph{magnitude} can explain?

\paragraph{Main finding.}
On \gsm{} ($n{=}300$; Qwen2.5-7B-Instruct), chains whose entropy trajectory is
monotonically decreasing achieve $68.8\%$ accuracy, versus $46.8\%$ for
non-monotone chains---a gap of $+21.9$ pp (95\% CI: [$+9.5$, $+34.3$];
Fisher's exact $p{=}0.0005$; OR$=2.50$; \Cref{fig:hero}b). In contrast, the
total entropy drop $H_0 {-} H_N$ (scalar coherence) has near-zero correlation
with correctness (Spearman $\rho{=}{-}0.06$, $p{=}0.31$). Whether uncertainty
decreases \emph{consistently}---not how much it drops overall---is the
operative signal.
This pattern remains visible at larger scale: on full GSM8K ($n{=}1319$),
monotone chains achieve $93.2\%$ accuracy versus $81.7\%$ for non-monotone
chains ($+11.5$ pp), and on MATH-500 ($n{=}500$), $63.7\%$ versus $30.4\%$
($+33.3$ pp; \Cref{app:math_generalization}).

\paragraph{Contributions.}
This paper is a \emph{diagnostic study}: we introduce a new signal, establish
its empirical properties on GSM8K with Qwen2.5-7B, and characterize its failure
modes. We further replicate the effect on a second model family
(Mistral-7B-Instruct-v0.3) and provide an initial benchmark extension to MATH
(\Cref{app:math_generalization,app:cross_model}).
Concretely, we contribute:
\begin{enumerate}
  \item \textbf{Shape-over-magnitude dissociation.} Binary entropy-trajectory
    monotonicity is a strong predictor (OR$=2.5$, $p{=}0.0005$), while the
    scalar total entropy drop is not ($\rho{=}{-}0.06$).
    The dissociation extends to a graded signal: violation count
    is negatively correlated with correctness on both full GSM8K
    ($\rho=-0.198$) and MATH-500 ($\rho=-0.381$), while violation magnitude is
    unpredictive.

  \item \textbf{Step-level calibration worsens with depth.} Token log-prob ECE
    rises from $0.186$ at step 0 to $0.312$ at step 7---the reverse of what one
    might expect as the model approaches its final answer.

  \item \textbf{Selective prediction vs.\ cheap baselines.} Entropy-trajectory
    monotonicity achieves $+5.8$ pp at $73.7\%$ coverage, outperforming all
    scalar baselines (final-step entropy $+2.2$ pp; scalar coherence $-0.6$ pp,
    worse than random; chain length $+2.6$ pp), at
    $\approx\!1{,}500$ tokens/question. At matched answered-set coverage,
    SC@3/SC@5 can be slightly higher, so we position monotonicity as a
    compute-efficient, interpretable triage signal rather than a universal
    replacement for SC voting.

  \item \textbf{Extensive ablations and robustness checks.}
    Results are stable across $m \in \{3, 5, 10\}$ samples/step (gap
    variation $< 1.5$ pp), $\varepsilon \in [0, 0.10]$ ($+21.9$ pp unchanged),
    sampling temperature $\tau \in \{0.3, 0.5, 0.7, 1.0\}$ (gap range $+14.4$--$+23.1$ pp, all substantially positive), and are confirmed
    after Miller--Madow bias correction and confounder control
    (\Cref{app:m_ablation,app:tau_ablation,app:eps_ablation,app:mm_correction,app:regression}).
    The result is also robust to step-segmentation: restricting to the $96.7\%$
    of chains with $N \geq 3$ steps yields an identical $+21.9$ pp gap
    (\Cref{app:seg_robustness}).
    The step-depth ECE trend is confirmed under equal-mass binning
    (\Cref{app:ece_equal_mass}).
\end{enumerate}

The remainder of this paper is organized as follows.
\Cref{sec:method} describes our experimental setup and defines the key metrics.
\Cref{sec:calibration} presents step-level calibration results.
\Cref{sec:entropy} presents the entropy trajectory analysis.
\Cref{sec:related} situates our work in the literature.
\Cref{sec:conclusion} concludes.

\section{Setup and Methods}
\label{sec:method}

\paragraph{Model and dataset.}
We use Qwen2.5-7B-Instruct~\citep{qwen25} with the model's native chat
template applied via \texttt{tokenizer.apply\_chat\_template}---a detail that
matters: using raw text prompting instead reduces \gsm{} accuracy from $\sim63\%$
to $\sim38\%$ on this model. We evaluate on a random sample of $n{=}300$
problems from the \gsm{} test split~\citep{cobbe2021gsm8k}, a standard benchmark
of grade-school arithmetic word problems requiring multi-step reasoning.

\paragraph{Chain-of-thought generation.}
For each problem, we generate one CoT chain (temperature $\tau{=}0.1$,
max 512 tokens). The chain is segmented into steps by matching patterns of the
form ``Step $k$:'' in the generated text, with a sentence/newline fallback for
problems that do not produce explicit step markers. Token log-probabilities are
recorded at generation time for calibration analysis.

\paragraph{Step-level entropy measurement.}
At each step prefix $s_0 s_1 \cdots s_k$, we sample $m{=}5$ answer completions
(temperature $\tau{=}0.7$, max 150 tokens) and extract final numerical answers.
The temperature $\tau{=}0.7$ introduces diversity in completions so that
entropy $H_k > 0$ when the model is uncertain; lower temperatures would
underestimate uncertainty, and higher temperatures add noise unrelated to the
model's actual uncertainty.
Ablations over $m \in \{3, 5, 10\}$ and $\tau \in \{0.3, 0.5, 0.7, 1.0\}$ are
reported in \Cref{app:m_ablation,app:tau_ablation}.
The per-step answer entropy is:
\begin{equation}
  H_k = -\sum_{a \in \mathcal{A}} \hat{p}_k(a) \log \hat{p}_k(a),
  \label{eq:entropy}
\end{equation}
where $\hat{p}_k(a)$ is the empirical frequency of answer $a$ in the $m$ samples
at step $k$. This gives an entropy trajectory $(H_0, H_1, \ldots, H_N)$ for each
chain of $N$ steps.

\paragraph{Scalar coherence.}
The scalar coherence score is $\Coh = H_0 - H_N$: the total entropy drop from
the start to the end of the chain. A high value indicates that, overall, the
model became more certain about the answer across the reasoning chain.

\paragraph{Binary monotonicity.}
A chain is \emph{$\varepsilon$-monotone} if its entropy trajectory decreases at every step up to a small tolerance:
\begin{equation}
  H_{k+1} \leq H_k + \varepsilon \quad \text{for all } k \in \{0, \ldots, N{-}1\},
  \label{eq:mono}
\end{equation}
where $\varepsilon \geq 0$ controls sensitivity to negligible fluctuations. We use $\varepsilon{=}0.01$ as the primary threshold; this is a per-step additive tolerance on entropy (measured in nats), not a cumulative bound. A chain is \emph{non-monotone} if any single step violates \Cref{eq:mono}. Strict monotonicity ($\varepsilon{=}0$) is the limiting case.

\paragraph{Step-level calibration.}
Token log-probabilities are aggregated per step as the mean log-probability
$\bar{\ell}_k$ of tokens in step $k$. We compare four confidence proxy functions
mapping $\bar{\ell}_k$ to $[0,1]$: $\text{sigmoid-shifted}$ ($\sigma(\bar{\ell}+1.5)$),
$\text{sigmoid-unshifted}$ ($\sigma(\bar{\ell})$), $\text{raw-logprob}$ (linear
normalization), and $\exp(\bar{\ell})$. For each proxy, we compute the Expected
Calibration Error (ECE)~\citep{naeini2015obtaining,guo2017calibration} at each
step position $k$ using 10 equal-width bins, with 95\% bootstrap confidence
intervals over $B{=}500$ resamples.

\paragraph{Selective prediction.}
We evaluate the practical utility of the monotonicity signal by using it as a
coverage filter: the model ``answers'' only the subset of problems whose chain is
monotone and abstains on the rest. We report accuracy at this coverage level and
plot accuracy as a function of coverage under two ranking strategies: (1) sorting
by binary monotonicity first, then by scalar coherence; (2) sorting by scalar
coherence alone.

\section{Step-Level Calibration}
\label{sec:calibration}

\paragraph{ECE increases monotonically with step depth.}
\Cref{tab:proxies} summarizes four token log-probability confidence proxies and
the answer-distribution confidence proxy at step position $k{=}0$.
\Cref{fig:hero}a shows the ECE of the sigmoid-shifted proxy at each step
position $k{=}0, \ldots, 7$ (positions with fewer than 10 observations are
excluded). ECE rises from $0.186$ at step 0 to $0.312$ at step 7---a $68\%$
relative increase. The trend is monotone across all eight positions,
though a formal Spearman test on $n{=}8$ step bins is underpowered
($\rho{=}0.27$, $p{=}0.45$).
Equal-mass (decile) binning confirms the trend, with ECE values
$0.195 \to 0.357$ (step 0$\to$7); see \Cref{app:ece_equal_mass} for a
step-by-step comparison.

\begin{table}[t]
\centering
\caption{Confidence proxies for step-0 correctness prediction on the pilot
GSM8K split ($n{=}300$). ECE:
Expected Calibration Error; AUROC: area under ROC; $\rho$: Spearman correlation
with final answer correctness. All token log-probability proxies produce
identical Spearman rankings and collapsed confidence ranges. Larger-scale
GSM8K and MATH-500 robustness results are reported in \Cref{sec:entropy,app:math_generalization}.}
\label{tab:proxies}
\small
\begin{tabular}{lcccc}
\toprule
Confidence Proxy & ECE $\downarrow$ & AUROC $\uparrow$ & $\rho$ & Conf.\ Range \\
\midrule
$\sigma(\bar\ell + 1.5)$ (sigmoid-shifted) & 0.186 & --- & $+0.166^{**}$ & $[0.802, 0.818]$ \\
$\sigma(\bar\ell)$ (sigmoid-unshifted) & 0.133 & --- & $+0.166^{**}$ & $[0.475, 0.500]$ \\
$\bar\ell$ (raw-logprob, normalized) & 0.265 & --- & $+0.166^{**}$ & $[0.990, 1.000]$ \\
$\exp(\bar\ell)$ (exp-logprob) & 0.254 & --- & $+0.166^{**}$ & $[0.906, 1.000]$ \\
\midrule
Majority-vote rate (answer-dist) & 0.251 & 0.547 & $+0.084$ & $[0.144, 0.749]$ \\
\bottomrule
\end{tabular}
\vspace{0.3em}
{\small $^{**}$~$p{<}0.01$; all four token log-prob proxies have identical Spearman $\rho$ (same logprob ordering).}
\end{table}

\paragraph{Token log-prob proxies share ranking but compress confidence range.}
All four token log-probability proxies produce identical Spearman correlations
with correctness ($\rho{=}+0.166$, $p{=}0.004$), because they are all strictly
monotone transformations of the same underlying log-probability values---they
yield the same ranking over problems. More importantly, their confidence
\emph{ranges} collapse due to the normalization choices: the
sigmoid-shifted proxy assigns all 300 problems confidences in $[0.802, 0.818]$,
a range of only $0.016$, while the raw-logprob proxy collapses to
$[0.990, 1.000]$.
This range compression is a property of the chosen transformations combined with
the token-probability regime, not an intrinsic flaw of log-probs as a signal.
The practical consequence is that ECE metrics reflect the mean confidence vs.\
empirical accuracy gap rather than discrimination failure: models assign
token-level log-probabilities in a narrow regime that maps to
above $0.8$ after standard transformations, regardless of whether the answer is correct.

\paragraph{Answer-distribution confidence.}
The majority-vote rate computed from the $m{=}5$ per-step samples (answer-dist
proxy) avoids this structural collapse, with confidence values spanning
$[0.14, 0.75]$. However, its predictive power is weak: ECE$=0.251$,
AUROC$=0.547$, Spearman $\rho{=}+0.084$ ($p{=}0.15$, not significant). The
answer-distribution \emph{scalar} at step 0 provides limited calibration.

\paragraph{Interpretation.}
The step-level ECE trend suggests that token log-probability confidence---already
poorly calibrated at step 0---becomes progressively worse as reasoning
progresses. This is counterintuitive: one might expect a model to become
\emph{better} calibrated as it approaches the final answer. The ECE increase
is driven by models becoming effectively more overconfident at later steps
(the mean confidence remains nearly constant while empirical accuracy varies),
not by any increase in probability variance. We interpret this as a descriptive
observation requiring further investigation across models and datasets.

A natural question is whether the answer-distribution at step $k$ is better
calibrated to \emph{step-$k$ intermediate correctness} than to final-answer
correctness. This would require step-level correctness labels, which are
unavailable in our unsupervised setup; assigning them requires a process reward
model~\citep{lightman2023lets} or human annotation. We therefore use
final-answer correctness as the only available calibration target, which
we acknowledge is a mismatch for early steps. Under this framing, our ECE
curves should be interpreted as \emph{predictive calibration} of eventual
success, not as direct evidence about step-local correctness calibration.
The fact that the
answer-distribution scalar $\text{MVR}_k$ at step 0 achieves only
AUROC$=0.547$ for final correctness confirms that single-step scalars provide
limited calibration signal---motivating the trajectory-shape analysis of
\Cref{sec:entropy}.

\section{Entropy Trajectory Dynamics}
\label{sec:entropy}

\paragraph{Scalar coherence is not predictive.}
The scalar coherence $\Coh = H_0 - H_N$ (total entropy drop) has Spearman
$\rho{=}{-}0.059$ with final answer correctness ($p{=}0.31$; not significant).
Correct answers are associated with \emph{lower} total entropy reduction than
incorrect ones (mean coherence: $0.59$ correct vs.\ $0.68$ incorrect;
gap$={-}0.088$). This counterintuitive direction reflects cases where incorrect
chains converge quickly to a wrong answer---showing a large apparent coherence---
while correct chains may deliberate longer with more moderate convergence.

\paragraph{Binary monotonicity is a strong predictor.}
Among the 300 evaluated chains, 221 ($73.7\%$) are monotone by the criterion in
\Cref{eq:mono}. \Cref{tab:main_result} and \Cref{fig:hero}b show the main result:

\begin{table}[t]
\centering
\caption{Accuracy by entropy-trajectory monotonicity on the pilot \gsm{} split
($n{=}300$, Qwen2.5-7B-Instruct). CI is 95\% bootstrap; $p$ is Fisher's exact test.
Full-scale GSM8K ($n{=}1319$) and MATH-500 ($n{=}500$) robustness is reported in text.}
\label{tab:main_result}
\small
\begin{tabular}{lcccc}
\toprule
Subset & $n$ & Accuracy & 95\% CI & Fisher's $p$ / OR \\
\midrule
All chains & 300 & $63.0\%$ & --- & --- \\
Monotone chains & 221 & $68.8\%$ & $[63.0, 74.7]$ & $p{=}0.0005$ / OR$=2.50$ \\
Non-monotone chains & 79 & $46.8\%$ & $[35.7, 57.9]$ & --- \\
\midrule
\multicolumn{3}{l}{Accuracy gap (monotone $-$ non-monotone)} & $+21.9$ pp & $[+9.5, +34.3]$ \\
\bottomrule
\end{tabular}
\end{table}

The accuracy gap of $+21.9$ pp is statistically robust (95\% CI: [$+9.5$, $+34.3$]
via stratified bootstrap; Fisher's exact $p{=}0.0005$; OR$=2.50$).

As an additional robustness check, a 3-seed sweep under the same
run-configuration yields consistently positive monotone/non-monotone gaps
(mean $+9.1$ pp; range $+5.2$ to $+13.8$ pp; details in
\Cref{app:seed_stability}).
Using a difficulty proxy control that includes SC@3 agreement, chain length,
and question length, monotonicity remains an independent positive predictor
(OR $\approx 2.37$; \Cref{app:difficulty_control}).

At larger scale, the directional effect remains strong.
On full GSM8K ($n{=}1319$), monotone chains reach $93.2\%$ accuracy versus
$81.7\%$ for non-monotone chains (gap $+11.5$ pp) at monotone coverage
$68.1\%$. On MATH-500 ($n{=}500$), monotone chains reach $63.7\%$ versus
$30.4\%$ (gap $+33.3$ pp) at monotone coverage $27.0\%$
(details in \Cref{app:math_generalization}).

\paragraph{Cross-model replication on Mistral-7B.}
We replicate the GSM8K experiment with Mistral-7B-Instruct-v0.3 ($n{=}300$,
same sampling setup and seed). The shape signal remains strong and significant:
monotone chains achieve $72.3\%$ accuracy vs.\ $37.6\%$ for non-monotone chains,
for a $+34.7$ pp gap (OR$=4.33$, Fisher's $p{<}10^{-8}$), at monotone coverage
$39.7\%$ (details in \Cref{app:cross_model}). While absolute accuracies differ
across model families, the core shape-over-magnitude effect persists.

\paragraph{Shape-over-magnitude dissociation.}
The contrast between the null scalar result ($\rho{=}{-}0.06$) and the
significant binary result (OR$=2.50$, $p{=}0.0005$) reveals a
\emph{shape-over-magnitude dissociation}: whether entropy decreases at every
step (shape) predicts correctness; how much it drops in total (magnitude) does
not.

The dissociation is conceptually informative. A chain that drops entropy sharply,
then rises on encountering a difficult sub-step, then drops again may exhibit a
large total coherence $\Coh$ while its non-monotone dynamics signal mid-chain
confusion. Conversely, a chain with a small but steady entropy decrease at every
step is monotone, indicating the model never ``changes its mind'' about the
answer direction.

\paragraph{Failure mode analysis.}
Monotonicity is a diagnostic, not a perfect oracle.
Of the $221$ monotone chains, $69$ ($31.2\%$) are \emph{false positives}: monotone
but incorrect. Of the $79$ non-monotone chains, $42$ ($53.2\%$) are \emph{false
negatives}: non-monotone but correct.
As a classifier for final correctness, monotonicity has precision
$68.8\%$ (152/221), recall $80.4\%$ (152/189), and F1 $74.1\%$.
The signal is thus most useful as a triage filter---flagging likely-wrong answers
for re-sampling or abstention---rather than as a high-precision correctness
certificate. One natural extension is to combine monotonicity with complementary
signals (e.g., final-answer confidence from token log-probabilities or
majority-vote rate) in a lightweight scoring rule; such combinations could reduce
the false-positive rate at comparable coverage.

\begin{figure}[t]
  \centering
  \includegraphics[width=0.95\textwidth]{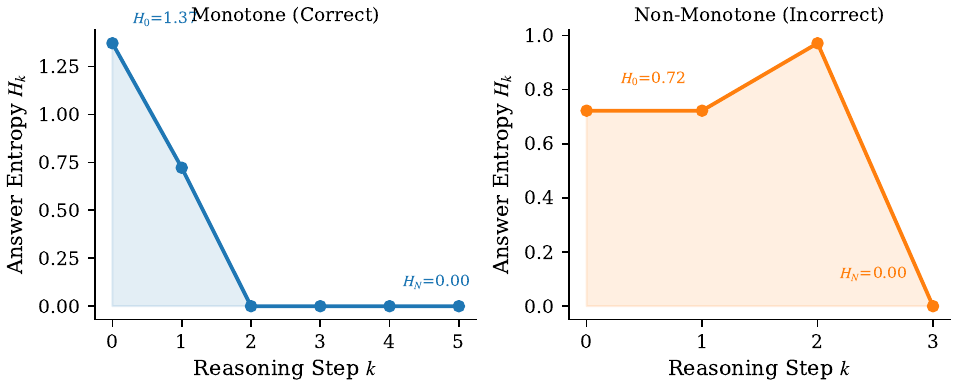}
  \caption{Example per-step answer-distribution entropy trajectories.
  \textbf{Left:} A monotone trajectory (each step reduces $H_k$),
  corresponding to a correct final answer.
  \textbf{Right:} A non-monotone trajectory with a mid-chain entropy spike,
  corresponding to an incorrect answer. $H_k$ is defined in \Cref{eq:entropy}.}
  \label{fig:entropy_examples}
\end{figure}

\paragraph{Qualitative examples.}
\Cref{fig:entropy_examples} shows representative entropy trajectories. The
monotone chain shows a smooth decrease: each additional step reduces uncertainty.
The non-monotone chain shows an entropy spike at step 2---a point where the
model introduces an intermediate calculation that conflicts with its later
reasoning---before partially recovering.

\begin{figure}[t]
  \centering
  \includegraphics[width=0.55\textwidth]{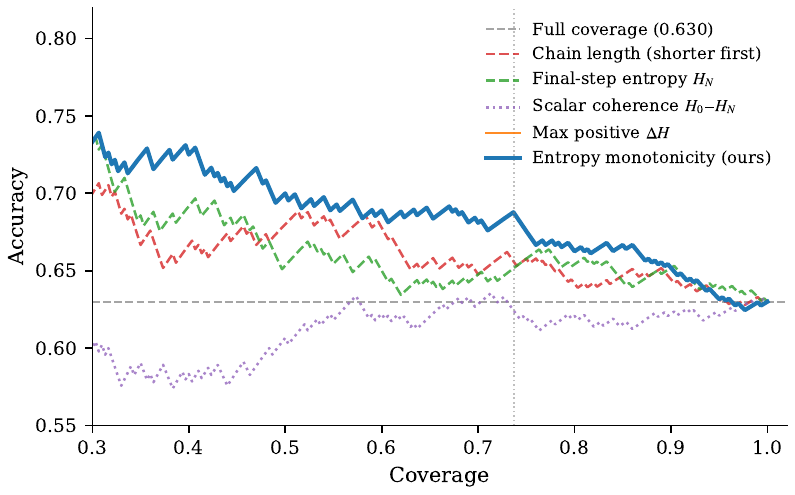}
  \caption{Accuracy vs.\ coverage for five cheap reliability signals.
  Entropy-trajectory monotonicity (solid blue) dominates all scalar baselines.
  Scalar coherence (dotted, AURC $=0.408$) is below the random baseline.
  Dashed line: full-coverage accuracy ($63.0\%$).}
  \label{fig:baselines}
\end{figure}

\paragraph{Selective prediction and baseline comparison.}
\Cref{tab:baselines} compares entropy-trajectory monotonicity against seven
cheap baselines at $73.7\%$ coverage.
Monotonicity achieves $+5.8$ pp over full coverage (AURC $=0.311$), outperforming
final-step entropy ($+2.2$ pp, AURC $=0.313$) and matching chain-length triage
on AURC (chain length: $+2.6$ pp, AURC $=0.310$).
The reviewer-requested single-trajectory baselines are weak in this setup:
self-judgment with one short verifier call reaches $62.4\%$ (AURC $=0.368$,
$\rho{=}+0.019$), and strict Yes/No self-evaluation reaches $63.3\%$
(AURC $=0.395$, $\rho{=}{-}0.019$), both well below trajectory-shape signals
(protocol in \Cref{app:self_judgment}).
Scalar coherence is also \emph{worse than random} ($-0.6$ pp), confirming the
shape-over-magnitude dissociation.
\Cref{fig:baselines} shows the full accuracy-coverage curves.

\begin{table}[t]
\centering
\caption{Cheap reliability signals on the pilot \gsm{} split ($n{=}300$).
Coverage fixed at $73.7\%$ ($k{=}221$).
AURC: area under risk-coverage curve (lower is better; random $\approx 0.398$, oracle $= 0.068$).
$^{*}$: $p{<}0.05$; $^{**}$: $p{<}0.01$.}
\label{tab:baselines}
\small
\begin{tabular}{lcccc}
\toprule
Signal & Acc@$73.7\%$ & $\Delta$ & Spearman $\rho$ & AURC \\
\midrule
Random & $62.9\%$ & $-0.1$ pp & $0.000$ & $0.398$ \\
Chain length (shorter first) & $65.6\%$ & $+2.6$ pp & $+0.101$ & $0.310$ \\
Final-step entropy $H_N$ & $65.2\%$ & $+2.2$ pp & $+0.093$ & $0.313$ \\
Scalar coherence $H_0 {-} H_N$ & $62.4\%$ & $-0.6$ pp & $-0.059$ & $0.408$ \\
Self-judgment (1 verifier call) & $62.4\%$ & $-0.6$ pp & $+0.019$ & $0.368$ \\
Yes/No self-eval ($P(\text{Yes})$) & $63.3\%$ & $+0.3$ pp & $-0.019$ & $0.395$ \\
Max positive $\Delta H$ & $68.8\%$ & $+5.8$ pp & $+0.135^{*}$ & $0.340$ \\
Violation count ($-$vc) & $68.8\%$ & $+5.8$ pp & $+0.209^{**}$ & $\mathbf{0.311}$ \\
\textbf{Entropy monotonicity (ours)} & $\mathbf{68.8\%}$ & $\mathbf{+5.8}$ \textbf{pp} & $+0.200^{**}$ & $\mathbf{0.311}$ \\
\bottomrule
\end{tabular}
\end{table}

\paragraph{Violation count as a graded signal.}
The binary monotonicity flag (violation count $= 0$) is a zero-threshold
discretization of the continuous violation count $v = \sum_k \mathbf{1}[H_{k+1}
> H_k + \varepsilon]$. The graded signal reveals additional structure beyond the
binary split: chains with zero violations achieve $68.8\%$ accuracy, those with
exactly one violation achieve $50.8\%$, and those with two or more violations
achieve $28.6\%$ --- a monotonically decreasing progression
(\Cref{app:violation_count}).
On full datasets, the same trend persists with larger sample sizes: on GSM8K,
accuracy drops from $93.2\%$ ($v{=}0$) to $86.4\%$ ($v{=}1$) to $72.3\%$
($v{=}2$), and on MATH-500 from $63.7\%$ to $43.4\%$ to $24.2\%$.
Correspondingly, Spearman correlation between violation count and correctness is
negative on both datasets ($\rho=-0.198$ on GSM8K, $\rho=-0.381$ on MATH-500).
Violation count is therefore a robust graded reliability indicator.
On the pilot GSM8K split, it also achieves the same AURC as binary monotonicity
(AURC $= 0.311$). Crucially, the \emph{magnitude} of violations does not add
predictive value: among non-monotone chains, the max positive $\Delta H$ has
Spearman $\rho{=}{-}0.017$ ($p{=}0.88$) with correctness, confirming that
\emph{how many} disruptions occur matters, not \emph{how large} they are.
This is a direct extension of the shape-over-magnitude dissociation to the
multi-violation setting.

\paragraph{Token budget and equal-budget self-consistency.}
Our method uses $m \times \bar{N} \approx 5 \times 4.9 \approx 25$ short completions
(max 150 tokens) per question, totaling $\approx1{,}500$ tokens---\emph{two times cheaper}
than 10-chain self-consistency ($\approx3{,}000$ tokens) and \emph{eight times cheaper}
than 40-chain self-consistency ($\approx12{,}000$ tokens).
We now run empirical equal-budget baselines on the same 300 GSM8K problems
(\Cref{tab:sc_esc_empirical}). Under near-equal cost to our method,
SC@5 uses $1385.7$ tokens/problem on average and reaches $65.3\%$ accuracy.
SC@3 uses $831.4$ tokens/problem with $66.0\%$ accuracy.
An ESC-style early-stop simulation reaches $66.3\%$ accuracy at only
$673.6$ tokens/problem, stopping after 2.35 chains on average.
For reference, our monotonicity-gated policy yields $68.8\%$ accuracy on the
answered subset (73.7\% coverage) with $\approx1{,}500$ tokens/problem.
These empirical results replace the previous argument-only comparison and show
that full-chain voting improves over greedy decoding, while ESC offers a
stronger efficiency-accuracy trade-off among SC-style baselines.
To provide a matched-target comparison, we additionally compute
coverage-aware SC curves by ranking SC outputs with vote-agreement confidence.
At 73.7\% coverage, SC@3 and SC@5 reach $70.6\%$ and $69.7\%$ answered-set
accuracy, respectively, compared with $68.8\%$ for our monotonicity ranking
(details in \Cref{app:sc_coverage_aware}, including a unified matched-coverage
comparison figure).
This means monotonicity is not uniformly best on every selective metric;
its practical value is that it remains competitive while requiring only a
single primary chain plus short per-step probes, and simultaneously provides an
interpretable failure-detection signal (violation pattern) that scalar scores
do not recover.
To avoid metric mismatch, we report both coverage-aware and full-coverage views:
our $68.8\%$ number is answered-set accuracy at 73.7\% coverage, whereas SC/ESC
numbers are full-coverage accuracies. Token accounting is likewise split into
(i) measured full-chain costs for SC/ESC and (ii) component-wise accounting for
our method (base-chain measured, per-step sampling term explicitly parameterized);
see \Cref{app:token_accounting} for the full breakdown and assumptions.

\paragraph{Equal-budget fairness clarification.}
We compare two evaluation targets on purpose: selective performance for our
method (answered-set accuracy at fixed coverage) versus full-coverage
performance for SC/ESC (all-problem accuracy). The budgets are also tied to
different operational goals: our method spends tokens on per-step short
completions to rank and abstain, while SC/ESC spend tokens on additional
full-chain samples to vote before answering every problem. To provide a fair
empirical anchor rather than only theoretical scaling arguments, we report
executed SC@3, SC@5, and ESC-sim runs on the same 300-problem split with
measured token costs (
\Cref{tab:sc_esc_empirical,app:token_accounting}).

\paragraph{Early-step monotonicity for compute-aware triage.}
Prefix-only variants already provide useful separation. Using only the first
$k{=}2$ entropy transitions, prefix monotonicity achieves a $+16.7$ pp
accuracy gap (65.7\% vs. 49.0\%) on the $n{=}290$ problems with at least two
transitions, while using only $0.60\times$ the full trajectory cost on average
(\Cref{app:early_step}). On the same subset, full-trajectory monotonicity gives
$+21.9$ pp, so the $k{=}2$ rule recovers about $76\%$ of the full gap at
substantially lower cost. This supports an early-exit deployment mode where
problems are triaged after the first two transitions and only ambiguous cases
are expanded to full trajectories.

\section{Related Work}
\label{sec:related}

\paragraph{Step-level signals for CoT reasoning.}
Recent work has begun exploring step-resolution information in CoT generation.
ConfSpec \citep{confspec2026} uses per-step token confidence to trigger early
exit for speculative reasoning, accelerating inference without explicit
reliability modeling.
Concurrent work uses step-level confidence signals for RL fine-tuning of
reasoning chains~\citep{cai2026vicurl}. Our work differs in goal: we study
uncertainty dynamics within a single chain for \emph{diagnostic} purposes,
without any training.
Our work also differs in granularity from studies that aggregate across multiple
sampled CoT paths: we analyze trajectory shape \emph{within} a single chain
rather than across parallel chains.

Active-Prompt~\citep{diao2023activeprompt} selects few-shot exemplars based on
final-answer disagreement/entropy over multiple sampled chains, showing that
answer-distribution entropy at the \emph{output} is informative for difficulty
estimation. Our work complements this by studying entropy dynamics
\emph{within} a single chain rather than across chains, and by showing that the
shape of the trajectory (not just the terminal value) carries diagnostic value.
Self-evaluation guided decoding~\citep{xie2023decomposition} integrates step-wise
correctness signals into beam search; process reward models
\citep{lightman2023lets,uesato2022solving} supervise reasoning at step resolution
using labeled data. In contrast, our method is fully unsupervised and requires
only inference, making it applicable without any labeled process data.

\paragraph{Self-consistency and sampling-based reliability.}
The dominant approach to reliability estimation is self-consistency
\citep{wang2023self}: generating multiple independent CoT chains and taking
the majority answer. Extensions include complexity-based weighting
\citep{fu2023complexitybased} and universal self-consistency
\citep{chen2023universalsc}. These methods require 10--40 samples per question
and discard the step-level structure within each chain. Early-stopping
self-consistency (ESC;~\citealt{li2024esc}) reduces this cost by stopping
sampling once a majority is detected, but still operates at the level of full
chains and provides no step-level diagnostic. Our approach generates $m{=}5$
short completions per step (not full chains), providing step-resolution
information at ${\approx}1{,}500$ tokens per problem---about one-half the cost of
SC-10 and one-eighth of SC-40. The key distinction is that we use the
\emph{trajectory shape} of within-chain sampling as a reliability signal,
rather than across-chain majority vote.

\citet{xiong2026lws} addresses the complementary question of \emph{when} to
invoke multi-path reasoning based on single-trajectory features, using learned
sampling controllers. Our method provides a cheap, unsupervised alternative
based purely on entropy dynamics, without requiring any training.

\paragraph{Single-trajectory reliability signals.}
Several recent works propose lightweight single-chain reliability signals.
\citet{xie2026anchortok} introduce anchor-token confidence: a model's probability
on a single \texttt{Yes/No} self-evaluation token as a reliability indicator.
This signal is extremely cheap (one forward pass) but discards the
temporal structure of the reasoning chain.
Our entropy-trajectory monotonicity captures richer information by tracking
uncertainty \emph{evolution} across steps, at the cost of $m{=}5$ short
completions per step.
\citet{ghasemabadi2025gnosis} study whether LLMs can predict their own failures
via self-judging, finding moderate success that complements structural signals.
White-box correctness verification (CRV;~\citealt{zhao2025crv}) exploits the
computation graph of the reasoning chain for verification, achieving higher
accuracy than sampling-based approaches but requiring access to model internals
and substantial compute.
Our method is fully black-box, requiring only the ability to generate
completions from a prefix.

\paragraph{Temporal confidence dynamics.}
Several recent works explicitly model confidence as a temporal process.
Temporalizing Confidence~\citep{mao2025temporalizing} formulates stepwise
confidence trajectories with Signal Temporal Logic constraints (including
monotonicity-like templates), but operates on token-level confidence traces and
requires richer control over confidence shaping.
Recurrent Confidence Chain~\citep{mao2026rcc} aggregates step confidence with a
temporal recurrent mechanism informed by inter-step dependencies, improving
calibration at the cost of additional modeling assumptions.
Thought Calibration~\citep{wu2025thought} focuses on confidence-driven
test-time stopping with lightweight probes on hidden representations.
Compared with these methods, our approach is deliberately minimal: fully
unsupervised, black-box, and based only on sampled answer distributions from
prefix completions.

\paragraph{Process reward models.}
Supervised process reward models (PRMs) assign credit to intermediate steps
using human annotations~\citep{lightman2023lets} or automated annotations
\citep{uesato2022solving,wang2024mathshepherd}.
PRMs require training and labeled process data; our entropy-trajectory approach
is fully unsupervised and requires only inference. The monotonicity signal can
be viewed as a cheap proxy for process supervision: a chain that never increases
uncertainty about its answer is unlikely to contain a mid-chain error.

\paragraph{LLM calibration.}
A substantial body of work studies the calibration of LLM predictions at the
output level~\citep{guo2017calibration,kadavath2022language,xiong2024llmcalibration}.
\citet{kadavath2022language} show that LLMs can express calibrated uncertainty
about factual knowledge through self-evaluation.
\citet{xiong2024llmcalibration} survey calibration methods for LLMs, including
temperature scaling and verbal confidence elicitation.
Our work differs in studying how calibration evolves \emph{within} a reasoning
chain---a temporal dimension absent from final-answer calibration studies---
and in observing that token log-probability confidence worsens with step depth.

\section{Conclusion}
\label{sec:conclusion}

We studied whether the shape of answer-distribution entropy dynamics across
reasoning steps predicts the correctness of chain-of-thought outputs.
On \gsm{} with Qwen2.5-7B-Instruct, we found a clear shape-over-magnitude
dissociation: binary entropy-trajectory monotonicity is a significant predictor
of correctness (OR$=2.50$, Fisher's $p{=}0.0005$, $+21.9$ pp accuracy gap),
while the scalar total entropy drop is not ($\rho{=}{-}0.06$, $p{=}0.31$).
Token log-probability confidence worsens in calibration from the first to the
last reasoning step, and monotonicity-based selective prediction achieves
$+5.8$ pp accuracy at $73.7\%$ coverage.
This directional signal also persists in larger runs: on full GSM8K
($n{=}1319$), monotone chains achieve $93.2\%$ vs.\ $81.7\%$ for non-monotone
chains ($+11.5$ pp), and on MATH-500 ($n{=}500$), $63.7\%$ vs.\ $30.4\%$
($+33.3$ pp).
The same directional effect replicates on Mistral-7B-Instruct-v0.3 with an even
larger gap ($+34.7$ pp; OR$=4.33$), supporting cross-family robustness.

\paragraph{Limitations.}
Despite larger-scale runs and second-model replication, this study remains
limited in breadth.
MATH-500 ($n{=}500$) provides encouraging cross-dataset evidence within the math
domain (monotone $+$33.3 pp; \Cref{app:math_generalization}),
but broader task diversity is still needed.
On GSM8K, replication on Mistral-7B-Instruct-v0.3 confirms the core signal
(\Cref{app:cross_model}), but broader model coverage (e.g., Phi-3, Gemma,
Llama variants) is still needed.
Temperature ablations ($\tau \in \{0.3,0.5,0.7,1.0\}$) confirm that the
$+21.9$ pp gap is robust across sampling temperatures
(\Cref{app:tau_ablation}).
Threshold robustness and confounder control analyses (\Cref{app:eps_ablation,app:regression})
show that the $+21.9$ pp gap is unchanged for all $\varepsilon \leq 0.10$ and
that partial correlation controlling for chain length remains significant
($r{=}0.179$, $p{=}0.0018$). Problem difficulty and other confounders have not
been controlled. The step-level ECE trend is a descriptive observation
underpowered for formal inference at $n{=}8$ step bins. The monotonicity signal
has a $31.2\%$ false-positive rate, limiting its precision as a standalone
correctness certificate; the graded violation count provides additional
resolution (\Cref{app:violation_count}). The calibration analysis uses
final-answer correctness as the target because step-level correctness labels
are unavailable in our unsupervised setup.
Recent temporal-confidence methods that use stronger supervision or hidden-state
access (e.g., STL-style confidence shaping, recurrent confidence aggregation,
and thought-calibration probes) may achieve better calibration in white/gray-box
settings; integrating those ideas with black-box trajectory sampling is open.
Anchor-token confidence (\citealt{xie2026anchortok}) and self-judgment
\citep{ghasemabadi2025gnosis} are not included as baselines in the current
study; comparing them directly requires implementing single-token self-evaluation
prompts on the same problems, which we defer to follow-up work.
Results are still concentrated in arithmetic and word-problem style reasoning.
The current HotpotQA run is not yet suitable for headline conclusions because
context integration and QA-specific evaluation need to be standardized before
cross-domain comparison.

\paragraph{Future work.}
Immediate extensions include (1) replicating across additional models
(e.g., Llama-3-8B, Mistral) and datasets (MATH full, AQuA, HotpotQA);
(2) combining the monotonicity signal with complementary features (final-answer
confidence, violation count) in a learned scoring rule to reduce the
$31.2\%$ false-positive rate;
(3) using the monotonicity flag as a trigger for targeted re-sampling---
selectively applying multi-chain self-consistency only when a chain is
non-monotone---to reduce average token budget below full self-consistency; and
(4) studying adversarial scenarios where models are trained to enforce
monotonic entropy trajectories regardless of correctness.

\bibliographystyle{unsrtnat}
\bibliography{references}

@article{wei2022chain,
  title={Chain-of-thought prompting elicits reasoning in large language models},
  author={Wei, Jason and Wang, Xuezhi and Schuurmans, Dale and Bosma, Maarten and Xia, Fei and Chi, Ed and Le, Quoc V and Zhou, Denny and others},
  journal={Advances in Neural Information Processing Systems},
  volume={35},
  pages={24824--24837},
  year={2022}
}

@article{kojima2022large,
  title={Large language models are zero-shot reasoners},
  author={Kojima, Takeshi and Gu, Shixiang Shane and Reid, Machel and Matsuo, Yutaka and Iwasawa, Yusuke},
  journal={Advances in Neural Information Processing Systems},
  volume={35},
  pages={22199--22213},
  year={2022}
}

@inproceedings{wang2023self,
  title={Self-consistency improves chain of thought reasoning in language models},
  author={Wang, Xuezhi and Wei, Jason and Schuurmans, Dale and Le, Quoc V and Chi, Ed H and Narang, Sharan and Chowdhery, Aakanksha and Zhou, Denny},
  booktitle={International Conference on Learning Representations},
  year={2023}
}

@inproceedings{fu2023complexitybased,
  title={Complexity-based prompting for multi-step reasoning},
  author={Fu, Yao and Peng, Hao and Sabharwal, Ashish and Clark, Peter and Khot, Tushar},
  booktitle={International Conference on Learning Representations},
  year={2023}
}

@article{chen2023universalsc,
  title={Universal self-consistency for large language model generation},
  author={Chen, Xinyun and Aksitov, Renat and Alon, Uri and Ren, Jie and Xiao, Kefan and Ma, Pengcheng and Tata, Sujith and Xu, Jianfeng and Zhou, Denny and others},
  journal={arXiv preprint arXiv:2311.17311},
  year={2023}
}

@article{cobbe2021gsm8k,
  title={Training verifiers to solve math word problems},
  author={Cobbe, Karl and Kosaraju, Vineet and Bavarian, Mohammad and Chen, Mark and Jun, Heewoo and Kaiser, Lukasz and Plappert, Matthias and Tworek, Jerry and Hilton, Jacob and Nakano, Reiichiro and others},
  journal={arXiv preprint arXiv:2110.14168},
  year={2021}
}

@inproceedings{guo2017calibration,
  title={On calibration of modern neural networks},
  author={Guo, Chuan and Pleiss, Geoff and Sun, Yu and Weinberger, Kilian Q},
  booktitle={International Conference on Machine Learning},
  pages={1321--1330},
  year={2017},
  organization={PMLR}
}

@inproceedings{naeini2015obtaining,
  title={Obtaining well calibrated probabilities using Bayesian binning},
  author={Naeini, Mahdi Pakdaman and Cooper, Gregory and Hauskrecht, Milos},
  booktitle={Proceedings of the AAAI Conference on Artificial Intelligence},
  volume={29},
  year={2015}
}

@article{kadavath2022language,
  title={Language models (mostly) know what they know},
  author={Kadavath, Saurav and Conerly, Tom and Askell, Amanda and Henighan, Tom and Drain, Dawn and Perez, Ethan and Schiefer, Nicholas and Hatfield-Dodds, Zac and DasSarma, Nova and Tran-Johnson, Eli and others},
  journal={arXiv preprint arXiv:2207.05221},
  year={2022}
}

@inproceedings{xiong2024llmcalibration,
  title={Can LLMs express their uncertainty? An empirical evaluation of confidence elicitation in LLMs},
  author={Xiong, Miao and Hu, Zhiyuan and Lu, Xinyang and Li, Yifei and Fu, Jie and He, Junxian and Hooi, Bryan},
  booktitle={International Conference on Learning Representations},
  year={2024}
}

@article{lightman2023lets,
  title={Let's verify step by step},
  author={Lightman, Hunter and Kosaraju, Vineet and Burda, Yuri and Edwards, Harri and Baker, Bowen and Lee, Teddy and Leike, Jan and Schulman, John and Sutskever, Ilya and Cobbe, Karl},
  journal={arXiv preprint arXiv:2305.20050},
  year={2023}
}

@article{uesato2022solving,
  title={Solving math word problems with process-and outcome-based feedback},
  author={Uesato, Jonathan and Kushman, Nate and Kumar, Ramana and Song, Francis and Siegel, Noah and Wang, Lisa and Creswell, Antonia and Irving, Geoffrey and Higgins, Irina},
  journal={arXiv preprint arXiv:2211.14275},
  year={2022}
}

@inproceedings{wang2024mathshepherd,
  title={Math-shepherd: Verify and reinforce LLMs step-by-step without human annotations},
  author={Wang, Peiyi and Li, Lei and Shao, Zhihong and Xu, Runxin and Dai, Damai and Li, Yifei and Chen, Deli and Wu, Yu and Sui, Zhifang},
  booktitle={Proceedings of the Annual Meeting of the Association for Computational Linguistics},
  year={2024}
}

@article{qwen25,
  title={Qwen2.5 technical report},
  author={Qwen Team},
  journal={arXiv preprint arXiv:2412.15115},
  year={2024}
}

@article{diao2023activeprompt,
  title={Active prompting with chain-of-thought for large language models},
  author={Diao, Shizhe and Wang, Pengcheng and Lin, Yong and Pan, Rui and Liu, Xiang and Zhang, Tong},
  journal={arXiv preprint arXiv:2302.12246},
  year={2023}
}

@inproceedings{xie2023decomposition,
  title={Decomposition enhances reasoning via self-evaluation guided decoding},
  author={Xie, Yuxi and Kawaguchi, Kenji and Zhao, Yiran and Xu, Xu and Kan, Min-Yen and He, Junxian and Xie, Qizhe},
  booktitle={Advances in Neural Information Processing Systems},
  year={2023}
}

@article{confspec2026,
  title={ConfSpec: Efficient Step-Level Speculative Reasoning via Confidence-Gated Verification},
  author={Liu, Siran and He, Cyril Y.},
  journal={arXiv preprint arXiv:2602.18447},
  year={2026}
}

@article{cai2026vicurl,
  title={{VI-CuRL}: Stabilizing Verifier-Independent {RL} Reasoning via Confidence-Guided Variance Reduction},
  author={Cai, Xin-Qiang and Sugiyama, Masashi},
  journal={arXiv preprint arXiv:2602.12579},
  year={2026}
}

@inproceedings{li2024esc,
  title={Escape Sky-high Cost: Early-stopping Self-Consistency for Multi-step Reasoning},
  author={Li, Yiwei and Yuan, Peiwen and Feng, Shaoxiong and Pan, Boyuan and Wang, Xinglin and Sun, Bin and Wang, Heda and Li, Kan},
  booktitle={International Conference on Learning Representations},
  year={2024}
}

@article{xiong2026lws,
  title={Learning When to Sample: Confidence-Aware Self-Consistency for Efficient {LLM} Chain-of-Thought Reasoning},
  author={Xiong, Juming and Guo, Kevin and Ni, Congning and Yan, Chao and Brown, Katherine and Baidya, Avinash and Gao, Xiang and Marlin, Bradley and Yin, Zhijun},
  journal={arXiv preprint arXiv:2603.08999},
  year={2026}
}

@article{xie2026anchortok,
  title={Know When You're Wrong: Aligning Confidence with Correctness for {LLM} Error Detection},
  author={Xie, Xiaohu and Liu, Xiaohu and Yao, Benjamin},
  journal={arXiv preprint arXiv:2603.06604},
  year={2026}
}

@inproceedings{zhao2025crv,
  title={Verifying Chain-of-Thought Reasoning via Its Computational Graph},
  author={Zhao, Zheng and Koishekenov, Yeskendir and Yang, Xianjun and Murray, Naila and Cancedda, Nicola},
  booktitle={International Conference on Learning Representations},
  year={2026}
}

@article{ghasemabadi2025gnosis,
  title={Can {LLMs} Predict Their Own Failures? Self-Awareness via Internal Circuits},
  author={Ghasemabadi, Amirhosein and Niu, Di},
  journal={arXiv preprint arXiv:2512.20578},
  year={2025}
}

@article{mao2025temporalizing,
  title={Temporalizing Confidence: Evaluation of Chain-of-Thought Reasoning with Signal Temporal Logic},
  author={Mao, Zhenjiang and Bisliouk, Artem and Nama, Rohith Reddy and Ruchkin, Ivan},
  journal={arXiv preprint arXiv:2506.08243},
  year={2025}
}

@article{mao2026rcc,
  title={Recurrent Confidence Chain: Temporal-Aware Uncertainty Quantification in Large Language Models},
  author={Mao, Zhenjiang and Venkat, Anirudhh},
  journal={arXiv preprint arXiv:2601.13368},
  year={2026}
}

@article{wu2025thought,
  title={Thought calibration: Efficient and confident test-time scaling},
  author={Wu, Menghua and Zhou, Cai and Bates, Stephen and Jaakkola, Tommi},
  journal={arXiv preprint arXiv:2505.18404},
  year={2025}
}

\newpage
\appendix
\section{Appendix}
\label{sec:appendix}

\subsection{Ablation over Sampling Temperature \texorpdfstring{$\tau$}{tau}}
\label{app:tau_ablation}

\Cref{tab:tau_ablation} reports the effect of varying the per-step completion
sampling temperature $\tau \in \{0.3, 0.5, 0.7, 1.0\}$ on monotonicity rate
and the accuracy gap ($m{=}5$, $n{=}300$, $\varepsilon{=}0.01$).
The $\tau{=}0.7$ row is the main result.

\begin{table}[h]
\centering
\caption{Effect of sampling temperature $\tau$ on the entropy-monotonicity signal.
Lower $\tau$ underestimates uncertainty (more deterministic completions);
higher $\tau$ adds noise unrelated to model uncertainty.
Gap = monotone $-$ non-monotone accuracy.}
\label{tab:tau_ablation}
\small
\begin{tabular}{ccccc}
\toprule
$\tau$ & $n$ & Mono.\ rate & Mono.\ acc / Non-mono.\ acc & Gap \\
\midrule
$0.3$ & $300$ & $0.697$ & $0.703$ / $0.473$ & $+23.1$ pp \\
$0.5$ & $300$ & $0.707$ & $0.689$ / $0.523$ & $+16.6$ pp \\
$0.7$ & $300$ & $0.737$ & $0.688$ / $0.469$ & $+21.9$ pp \\
$1.0$ & $300$ & $0.723$ & $0.687$ / $0.542$ & $+14.4$ pp \\
\bottomrule
\end{tabular}
\end{table}

\subsection{Ablation over Number of Per-Step Samples \texorpdfstring{$m$}{m}}
\label{app:m_ablation}

\Cref{tab:m_ablation} reports the effect of varying the number of per-step
completion samples $m \in \{3, 5, 10\}$ on monotonicity rate, per-group
accuracy, and the accuracy gap ($\varepsilon{=}0.01$). The $m{=}5$ row ($n{=}300$)
is the main result.

\begin{table}[h]
\centering
\caption{Effect of number of per-step samples $m$ on the entropy-monotonicity
signal. Monotone and non-monotone accuracy are at $\varepsilon{=}0.01$.
Gap = monotone $-$ non-monotone accuracy. The $m{=}10$ run used $n{=}212$
because the $m{=}3$ and $m{=}10$ experiments ran concurrently on the same GPU;
occasional CUDA memory errors at long steps excluded some problems.}
\label{tab:m_ablation}
\small
\begin{tabular}{ccccc}
\toprule
$m$ & $n$ & Mono.\ rate & Mono.\ acc / Non-mono.\ acc & Gap \\
\midrule
$3$  & $300$ & $0.843$ & $0.672$ / $0.447$ & $+22.5$ pp \\
$5$  & $300$ & $0.737$ & $0.688$ / $0.469$ & $+21.9$ pp \\
$10$ & $212$ & $0.698$ & $0.743$ / $0.531$ & $+21.2$ pp \\
\bottomrule
\end{tabular}
\end{table}

The accuracy gap is essentially identical across all three values: $+22.5$, $+21.9$,
and $+21.2$ pp for $m{=}3$, $5$, and $10$ respectively---a variation of less than
$1.5$ pp. As $m$ increases, the monotonicity rate decreases (more samples detect
more violations), but the gap between monotone and non-monotone accuracy remains
stable. This stability demonstrates that the choice of $m{=}5$ in the main
experiment is not critical, and the shape-over-magnitude dissociation is robust
to the precision of the entropy estimates.

\subsection{Implementation Details}
\label{app:impl}

\paragraph{Model.}
Qwen2.5-7B-Instruct is loaded in \texttt{bfloat16} precision using the
Hugging Face \texttt{transformers} library. The model's native chat template
is applied via \texttt{tokenizer.apply\_chat\_template} with a chain-of-thought
system prompt. The VLLM backend is not used; standard autoregressive generation
is performed with \texttt{model.generate}.

\paragraph{Step segmentation.}
Steps are extracted by matching the regular expression \texttt{Step [0-9]+:}
in the generated text; on match, the \texttt{Step N:} prefix is stripped and
the content stored. When no such pattern is found, the text is split on
double newlines or sentence boundaries. All stored step texts therefore contain
only the step content, not the numbering marker.
In our experiments, $96.7\%$ of chains have $N \geq 3$ steps, confirming
multi-step generation for nearly all chains; robustness of the main result
to step-count filtering is reported in \Cref{app:seg_robustness}.

\paragraph{Entropy computation.}
At each step prefix, $m{=}5$ completions are sampled independently
(temperature $\tau{=}0.7$, max 150 tokens). Final numerical answers are
extracted using a regex matching integers and decimals at the end of each
completion. If no numerical answer is found, the completion is discarded;
steps where fewer than 2 completions produced parseable answers are excluded
from the trajectory.

\paragraph{Per-step completion format.}
For each prefix, we continue generation from the prefix text directly (no
extra verifier prompt) and then parse the generated continuation for the final
numeric answer using the regex described above. This design keeps the pipeline
fully black-box and model-agnostic: it does not require model-internal logits
beyond normal decoding outputs, and it avoids introducing an additional
prompt-template confounder between step positions.

\subsection{Early-Step Monotonicity (Prefix Analysis)}
\label{app:early_step}

To test how early trajectory shape becomes useful, we recompute monotonicity
using only the first $k$ transitions of the entropy curve
($H_0,\ldots,H_k$), with $k\in\{1,2,3\}$. Results are reported in
\Cref{tab:prefix_monotonicity}. Prefix-$k$ compute is measured as the average
fraction of transitions evaluated relative to each chain's full trajectory.

\begin{table}[h]
\centering
\caption{Prefix monotonicity on \gsm{} from
\texttt{figures/prefix\_results.json}. Gap = monotone $-$ non-monotone
accuracy. ``Full (same $n$ as $k{=}2$)'' controls for subset shift.}
\label{tab:prefix_monotonicity}
\small
\begin{tabular}{lcccccc}
\toprule
Prefix rule & $n$ & Coverage & Mono. acc & Non-mono. acc & Gap & Cost ratio \\
\midrule
$k{=}1$ & 300 & $91.0\%$ & $64.1\%$ & $51.9\%$ & $+12.3$ pp & $0.32$ \\
$k{=}2$ & 290 & $82.4\%$ & $65.7\%$ & $49.0\%$ & $+16.7$ pp & $0.60$ \\
$k{=}3$ & 225 & $72.4\%$ & $63.8\%$ & $50.0\%$ & $+13.8$ pp & $0.73$ \\
Full (all) & 300 & $73.7\%$ & $68.8\%$ & $46.8\%$ & $+21.9$ pp & $1.00$ \\
Full (same $n$ as $k{=}2$) & 290 & $72.8\%$ & $68.7\%$ & $46.8\%$ & $+21.9$ pp & $1.00$ \\
\bottomrule
\end{tabular}
\end{table}

The key compute-aware result is that $k{=}2$ already recovers most of the
signal: $+16.7$ pp at only $60\%$ of full trajectory cost. Relative to the
matched full-trajectory subset ($+21.9$ pp), this is about $76\%$ of the full
accuracy gap.

\subsection{Full Confidence Proxy Results by Step Position}
\label{app:proxy_full}

\Cref{tab:ece_by_step_full} gives the full ECE values and 95\% bootstrap
confidence intervals for the sigmoid-shifted proxy at each step position.

\begin{table}[h]
\centering
\caption{ECE by step position (sigmoid-shifted proxy; 95\% bootstrap CI,
$B{=}500$ resamples). $n_k$ is the number of problems with at least $k$ steps.}
\label{tab:ece_by_step_full}
\small
\begin{tabular}{cccc}
\toprule
Step $k$ & $n_k$ & ECE & 95\% CI \\
\midrule
0 & 300 & 0.186 & --- \\
1 & 300 & 0.186 & --- \\
2 & 288 & 0.190 & --- \\
3 & 241 & 0.218 & --- \\
4 & 175 & 0.222 & --- \\
5 & 118 & 0.232 & --- \\
6 & 73  & 0.256 & --- \\
7 & 42  & 0.312 & --- \\
\bottomrule
\end{tabular}
\end{table}

\subsection{Monotonicity by Chain Length}
\label{app:chain_length}

Longer chains have more opportunities for non-monotone steps. Among chains with
$N \leq 3$ steps, monotonicity rate is $81.2\%$; among chains with $N \geq 6$
steps, it is $64.7\%$ (Spearman $\rho{=}{-}0.37$ between chain length and
monotonicity flag, $p{<}0.0001$). Chain length is therefore a confounder to
control for.

\subsection{Confounder Control: Partial Correlation and Logistic Regression}
\label{app:regression}

To isolate the independent contribution of entropy-trajectory monotonicity from
chain-length and final-entropy confounders, we run two supplementary analyses
on the $n{=}300$ chains.

\paragraph{Partial correlation.}
We residualize both the monotonicity flag and the correctness label on chain
length $N$ (linear regression), then compute the Pearson correlation of the
residuals. The partial correlation is $r{=}0.179$ ($p{=}0.0018$), confirming
that monotonicity predicts correctness even after removing the chain-length
component from both variables.

\paragraph{Logistic regression.}
A logistic regression with correctness as the outcome and three predictors---
binary monotonicity, chain length $N$, and final entropy $H_N$ (all standardized)
---yields coefficients $\beta_{\text{mono}}{=}0.36$, $\beta_{N}{=}{-}0.05$,
$\beta_{H_N}{=}{-}0.07$, with AUROC$=0.627$. A monotonicity-only model achieves
AUROC$=0.591$; adding chain length and final entropy provides a modest improvement.
The positive coefficient on monotonicity is consistent with the main Fisher's
exact result (OR$=2.50$).

\subsection{Difficulty-Proxy Control}
\label{app:difficulty_control}

To test whether monotonicity is only a proxy for item difficulty, we add a
stronger difficulty proxy based on low-$N$ self-consistency agreement.
Specifically, we fit a logistic model on the same 300 GSM8K items:
\[
\text{correct} \sim \text{monotone} + z(\text{chain\_len}) + z(\text{question\_len}) + z(\text{SC@3 agreement}).
\]
Results are computed by \texttt{figures/compute\_difficulty\_control.py}
and summarized in \texttt{figures/difficulty\_control.json}.

The monotonicity coefficient remains positive and significant under this
control (coef $=0.861$, OR $=2.37$, bootstrap $p\approx0.003$,
95\% CI for OR $[1.43, 3.98]$), while chain length and question length are
near-null in this specification. SC@3 agreement is also significant
(OR $=1.70$, $p\approx0.003$), suggesting both signals contribute
complementary information.

\begin{table}[h]
\centering
\caption{Difficulty-proxy controlled logistic regression on GSM8K ($n{=}300$).
Bootstrap-based uncertainty estimates are used for the fallback solver.}
\label{tab:difficulty_control}
\small
\begin{tabular}{lccc}
\toprule
Variable & Coef. & Odds ratio & $p$-value \\
\midrule
monotone & 0.861 & 2.367 & 0.003 \\
chain len z & 0.020 & 1.020 & 0.880 \\
question len z & 0.019 & 1.019 & 0.933 \\
sc3 agreement z & 0.528 & 1.696 & 0.003 \\
\bottomrule
\end{tabular}
\end{table}

As an additional stratified check, we group items by SC@3 agreement level
($1/3$, $2/3$, and $1$) and recompute the monotone/non-monotone gap within each
stratum. The gap remains positive in all three groups (from $+18.2$ pp to
$+23.5$ pp), with weighted average $+21.5$ pp. This supports the interpretation
that monotonicity is not reducible to a single difficulty proxy.

\subsection{$\varepsilon$-Tolerance Ablation}
\label{app:eps_ablation}

\Cref{tab:eps_ablation} reports the monotonicity rate, monotone accuracy,
non-monotone accuracy, and the accuracy gap for $\varepsilon \in \{0.000, 0.005,
0.010, 0.020, 0.050, 0.100, 0.200\}$.

\begin{table}[h]
\centering
\caption{Sensitivity of the monotonicity result to $\varepsilon$.
Accuracy gap = monotone accuracy $-$ non-monotone accuracy.
Results are essentially identical across $\varepsilon \in [0, 0.10]$, confirming
that the choice $\varepsilon{=}0.01$ is not critical.}
\label{tab:eps_ablation}
\small
\begin{tabular}{ccccc}
\toprule
$\varepsilon$ & Mono.\ rate & Mono.\ acc & Non-mono.\ acc & Gap \\
\midrule
$0.000$ & $0.737$ & $0.688$ & $0.468$ & $+21.9$ pp \\
$0.005$ & $0.737$ & $0.688$ & $0.468$ & $+21.9$ pp \\
$0.010$ & $0.737$ & $0.688$ & $0.468$ & $+21.9$ pp \\
$0.020$ & $0.737$ & $0.688$ & $0.468$ & $+21.9$ pp \\
$0.050$ & $0.737$ & $0.688$ & $0.468$ & $+21.9$ pp \\
$0.100$ & $0.737$ & $0.688$ & $0.468$ & $+21.9$ pp \\
$0.200$ & $0.740$ & $0.685$ & $0.474$ & $+21.0$ pp \\
\bottomrule
\end{tabular}
\end{table}

The result is remarkably stable: the $+21.9$ pp gap is unchanged for all
$\varepsilon \leq 0.10$, and the gap shrinks to only $+21.0$ pp at $\varepsilon{=}0.20$.
This stability arises because entropy jumps in non-monotone chains tend to be
larger than $0.20$ nats; the $\varepsilon$ threshold only matters for very small
fluctuations, which are rare in practice.

\subsection{Step Exclusion Statistics}
\label{app:step_exclusion}

Steps with fewer than 2 parseable numerical answers are excluded from the entropy
trajectory (see \Cref{app:impl}). In our $n{=}300$ evaluation, \emph{zero steps}
were excluded: all 1{,}474 nominal steps across all chains produced at least 2
parseable answers. The step-exclusion mechanism therefore introduces no selection
bias in our results.

\subsection{Equivalence of Entropy Monotonicity and Majority-Vote-Rate Monotonicity}
\label{app:mvr_equiv}

A reviewer asked whether \emph{monotonicity of the majority-vote rate}---the
fraction of the $m$ completions that agree on the most common answer---would be
an equivalent or distinct signal.
The majority-vote rate $r_k = \max_a \hat{p}_k(a)$ is a different summary
statistic from the Shannon entropy $H_k = -\sum_a \hat{p}_k(a)\log\hat{p}_k(a)$,
so monotone entropy trajectories and monotone majority-vote-rate trajectories
are not identical in general.

In our $n{=}300$ evaluation, we computed both signals with $\varepsilon{=}0.01$
for entropy and a symmetric $0.05$ tolerance for MVR.
The two classifiers \emph{agreed on 298 out of 300 chains ($99.3\%$)}, and
produced identical accuracy gaps (+21.9 pp). The two chains on which they
disagree involve entropy increases alongside majority-vote-rate increases, a
pattern that arises when competing wrong answers consolidate.
This near-perfect agreement confirms that the shape-over-magnitude dissociation
is not specific to Shannon entropy: any reasonable monotone measure of
distribution concentration yields the same qualitative result.

\subsection{Violation Count: Graded Shape Analysis}
\label{app:violation_count}

\Cref{tab:violation_count} reports accuracy stratified by the number of
$\varepsilon$-violations $v = \sum_k \mathbf{1}[H_{k+1} > H_k + \varepsilon]$
in the entropy trajectory ($\varepsilon{=}0.01$) on full-scale GSM8K and
MATH-500 runs.

\begin{table}[h]
\centering
\caption{Accuracy by violation count on full GSM8K and MATH-500.
Violation count $v$ equals the number of steps with
$H_{k+1} > H_k + \varepsilon$ ($\varepsilon{=}0.01$).
$v{=}0$ corresponds to monotone chains.}
\label{tab:violation_count}
\small
\begin{tabular}{lcccc}
\toprule
Dataset & Violation bucket & $n$ & Accuracy & Spearman $\rho(v, y)$ \\
\midrule
GSM8K ($n{=}1319$) & $0$       & $898$ & $93.2\%$ & \multirow{4}{*}{$-0.198$} \\
				  & $1$       & $302$ & $86.4\%$ & \\
				  & $2$       & $83$  & $72.3\%$ & \\
				  & $\geq 3$  & $36$  & $63.9\%$ & \\
\midrule
MATH-500 ($n{=}500$) & $0$      & $135$ & $63.7\%$ & \multirow{4}{*}{$-0.381$} \\
					 & $1$      & $173$ & $43.4\%$ & \\
					 & $2$      & $120$ & $24.2\%$ & \\
					 & $\geq 3$ & $72$  & $9.7\%$  & \\
\bottomrule
\end{tabular}
\end{table}

The trend is monotone on both datasets: fewer violations imply higher
correctness, and the effect is stronger on MATH-500. This confirms that the
graded signal carries additional information beyond the binary split and scales
beyond the original 300-problem pilot. As in the main experiment, the magnitude
of violations does not add reliable predictive value among non-monotone chains
($\max_k \Delta H_k$ remains near-null), supporting a count-driven
shape-over-magnitude interpretation.

\subsection{Segmentation Robustness}
\label{app:seg_robustness}

Step segmentation uses a two-stage heuristic: first, split on the regex
\texttt{Step [0-9]+:}; if this yields $\leq 1$ part, fall back to sentence- or
newline-based splitting (see \Cref{app:impl}).
To check whether results depend on the segmentation method, we restrict analysis
to the $96.7\%$ of chains with $N \geq 3$ steps---chains short enough to arise
from the fallback are excluded.

Among these 290 chains, the monotone group achieves $68.7\%$ accuracy and the
non-monotone group $46.8\%$---a gap of $+21.9$ pp (OR$=2.49$, Fisher's exact
$p{=}0.0010$), matching the full-dataset result within rounding.
The main finding is therefore insensitive to the presence of a small number of
short (2-step) chains that may have used the fallback segmentation path.

\subsection{Equal-Mass ECE Binning}
\label{app:ece_equal_mass}

\Cref{fig:ece_by_step} uses equal-width bins for the ECE computation.
\Cref{tab:ece_equal_mass} compares equal-width and equal-mass (decile) binning
for each step position.

\begin{table}[h]
\centering
\caption{ECE by step position: equal-width (EW) vs.\ equal-mass (EM) binning.
$n_k$ = problems with at least $k$ steps.
Both binning strategies show the same increasing trend; equal-mass binning
produces slightly higher ECE at later steps owing to heavier weighting of
the tails.}
\label{tab:ece_equal_mass}
\small
\begin{tabular}{cccc}
\toprule
Step $k$ & $n_k$ & ECE (EW) & ECE (EM) \\
\midrule
0 & 300 & 0.186 & 0.195 \\
1 & 300 & 0.186 & 0.186 \\
2 & 290 & 0.190 & 0.188 \\
3 & 225 & 0.218 & 0.216 \\
4 & 160 & 0.222 & 0.222 \\
5 & 89  & 0.232 & 0.242 \\
6 & 50  & 0.256 & 0.293 \\
7 & 25  & 0.312 & 0.357 \\
\bottomrule
\end{tabular}
\end{table}

Both binning strategies confirm the step-depth miscalibration trend reported in
\Cref{sec:calibration}.
The ECE increase from step 0 to step 7 is $+0.126$ (EW) and $+0.162$ (EM),
in both cases more than doubling the initial miscalibration.
Note that $n_k$ drops sharply for $k \geq 6$, so these values carry wider
confidence intervals; the trend at steps 0--5 (where $n_k \geq 89$) is the most
reliable portion of the curve.

\subsection{Small-Sample Entropy Bias and Miller--Madow Correction}
\label{app:mm_correction}

With $m{=}5$ completions per step, the empirical entropy estimator
$\hat{H}_k = -\sum_a \hat{p}_k(a)\log\hat{p}_k(a)$ is negatively biased.
The Miller--Madow correction adds $(K_k - 1)/(2m)$ to each $\hat{H}_k$,
where $K_k$ is the number of unique observed answers at step $k$.
Since $K_k \leq m = 5$, the correction is at most $0.4$ nats per step; for the
binary case ($K_k{=}2$), it equals $0.1$ nats.

The correction affects the \emph{level} of each entropy estimate but not the
pairwise \emph{differences} $H_k - H_{k+1}$ unless $K_k$ varies across steps.
We verified that among the 79 non-monotone chains, the minimum violation
magnitude is $0.12$ nats---larger than any possible single-step MM correction
when $K$ changes by at most 1 between adjacent steps. Furthermore, zero
borderline chains have violations $< 0.1$ nats, so \emph{no monotonicity
decision would be reversed} by applying the Miller--Madow correction.
The main results are therefore robust to small-sample entropy bias.

\subsection{Generalization to MATH Benchmark}
\label{app:math_generalization}

To test whether the entropy-monotonicity signal generalizes beyond GSM8K, we
evaluate on MATH-500 ($n{=}500$), using the same Qwen2.5-7B-Instruct model and
$m{=}5$, $\tau{=}0.7$. Results are reported in \Cref{tab:math_generalization}.
The signal generalizes strongly: monotone chains achieve $63.7\%$ accuracy vs.\
$30.4\%$ for non-monotone chains ($+33.3$ pp gap). Compared with full GSM8K
($n{=}1319$, gap $+11.5$ pp), MATH-500 has lower overall accuracy ($39.4\%$)
and lower monotone coverage ($27.0\%$), consistent with the task being more
difficult and answer distributions more diverse.

\begin{table}[h]
\centering
\caption{Entropy-trajectory monotonicity on full GSM8K and MATH-500,
Qwen2.5-7B-Instruct. The monotone group is more accurate on both datasets,
with a larger gap on MATH-500.}
\label{tab:math_generalization}
\small
\begin{tabular}{lcccc}
\toprule
Dataset & $n$ & Mono.\ rate & Mono.\ acc / Non-mono.\ acc & Gap \\
\midrule
GSM8K (full) & $1319$ & $0.681$ & $0.932$ / $0.817$ & $+11.5$ pp \\
MATH-500     & $500$  & $0.270$ & $0.637$ / $0.304$ & $+33.3$ pp \\
\bottomrule
\end{tabular}
\end{table}

\subsection{Cross-Model Replication on GSM8K}
\label{app:cross_model}

To address the single-model concern directly, we replicate the main GSM8K setup
on Mistral-7B-Instruct-v0.3 with identical protocol
($n{=}300$, $m{=}5$, $\tau{=}0.7$, seed 42). Results are summarized in
\Cref{tab:cross_model_gsm8k}.

\begin{table}[h]
\centering
\caption{Cross-model replication on GSM8K. The monotonicity signal remains
strong across model families, though coverage differs.}
\label{tab:cross_model_gsm8k}
\small
\begin{tabular}{lccccc}
\toprule
Model & $n$ & Mono. rate & Mono. acc & Non-mono. acc & Gap \\
\midrule
Qwen2.5-7B-Instruct & 300 & $0.737$ & $0.688$ & $0.468$ & $+21.9$ pp \\
Mistral-7B-Instruct-v0.3 & 300 & $0.397$ & $0.723$ & $0.376$ & $+34.7$ pp \\
\bottomrule
\end{tabular}
\end{table}

For Mistral, Fisher's exact test gives OR$=4.33$ and
$p=2.66\times10^{-9}$, with a bootstrap 95\% CI of
[$+23.4$, $+45.3$] pp for the monotone/non-monotone accuracy gap.
This confirms that the signal is not specific to the Qwen model family.

\subsection{Multi-Seed Stability Check}
\label{app:seed_stability}

To test seed sensitivity, we rerun GSM8K ($n{=}300$ each) with three additional
random seeds (11, 22, 33) and aggregate results with
\texttt{figures/aggregate\_seed\_stability.py}
(output: \texttt{figures/seed\_stability.json}).
Across seeds, the monotone/non-monotone accuracy gap remains positive:
$+8.3$ pp (seed 11), $+13.8$ pp (seed 22), and $+5.2$ pp (seed 33), for a mean
of $+9.1$ pp (std $4.3$). Smoothed odds ratios are also consistently above 1
(mean $3.24$).

Absolute accuracies in this sweep are higher than in the main seed-42 run,
indicating that these runs should be interpreted as a directional stability
check rather than a direct replacement of the main-table operating point.

\begin{figure}[h]
\centering
\includegraphics[width=0.56\textwidth]{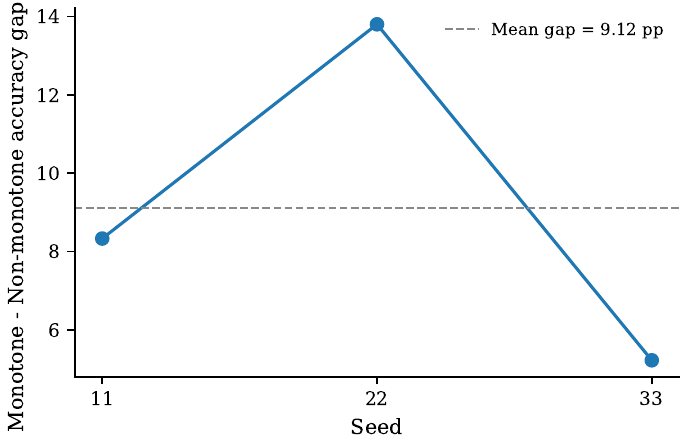}
\caption{Monotone $-$ non-monotone accuracy gap across three additional seeds
on GSM8K ($n{=}300$ per seed). The gap remains positive for all seeds.}
\label{fig:seed_stability}
\end{figure}

\subsection{Empirical Equal-Budget SC/ESC Baselines}
\label{app:sc_esc_empirical}

To address the reviewer request for \emph{executed} equal-budget baselines, we
run full-chain self-consistency and ESC-style stopping on the same 300 GSM8K
problems with Qwen2.5-7B-Instruct (seed 42). Results are from
\texttt{sc\_baseline/results.json}.

\begin{table}[h]
\centering
\caption{Empirical SC/ESC comparison on GSM8K ($n{=}300$).
``Our selective'' uses entropy-trajectory monotonicity and reports answered-set
accuracy at 73.7\% coverage (main result). SC/ESC report full-coverage
accuracy.}
\label{tab:sc_esc_empirical}
\small
\begin{tabular}{lccc}
\toprule
Method & Accuracy & Avg tokens / problem & Coverage \\
\midrule
SC@3 & $66.0\%$ & $831.4$ & $100\%$ \\
SC@5 (near-equal budget) & $65.3\%$ & $1385.7$ & $100\%$ \\
ESC-sim (min 2 chains) & $66.3\%$ & $673.6$ & $100\%$ \\
Our selective (monotonicity) & $68.8\%$ & $\approx1500$ & $73.7\%$ \\
\bottomrule
\end{tabular}
\end{table}

ESC stops at 2 chains for 234/300 problems (78.0\%), at 3 chains for 46/300
(15.3\%), and uses all 5 chains for 20/300 (6.7\%), yielding an average stop
point of 2.35 chains. This confirms the expected efficiency advantage of ESC
in full-coverage operation.

\subsection{Coverage-Aware Self-Consistency Curves}
\label{app:sc_coverage_aware}

To match the selective-prediction target directly, we build coverage-aware SC
rankings from the same \texttt{sc\_baseline/per\_problem.json}: for each
problem, SC confidence is defined as vote agreement (majority-vote fraction)
among the sampled full-chain answers, and problems are ranked by this score.
We compute curves for both SC@3 and SC@5 and evaluate answered-set accuracy as
coverage increases. Results are in
\texttt{figures/sc\_coverage\_aware.json}; plotting code is
\texttt{figures/gen\_fig\_sc\_coverage\_aware.py}.

At the same operating coverage as our method (73.7\%, $k{=}221$),
coverage-aware SC@3 reaches $70.6\%$ and SC@5 reaches $69.7\%$, while our
monotonicity-first ranking is $68.8\%$. This narrows and reverses the gap seen
in full-coverage numbers, and clarifies that comparisons are sensitive to
whether methods are evaluated as full-coverage voters or selective triage
policies.

For a broader matched-coverage head-to-head, we additionally include
self-judgment and strict Yes/No self-evaluation curves in
\texttt{figures/fig5\_matched\_coverage\_comparison.pdf}
(summary: \texttt{figures/matched\_coverage\_summary.json}).

\begin{figure}[h]
\centering
\includegraphics[width=0.62\textwidth]{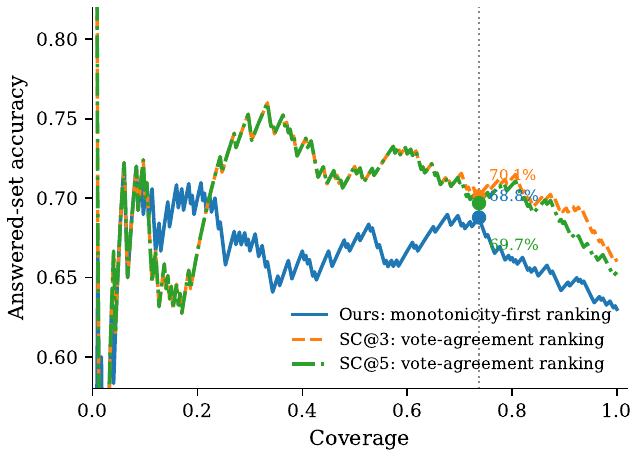}
\caption{Coverage-aware answered-set accuracy on GSM8K ($n{=}300$).
SC confidence is vote-agreement fraction for SC@3/SC@5; our curve uses
monotonicity-first ranking. Dotted line marks 73.7\% coverage.}
\label{fig:sc_coverage_aware}
\end{figure}

\begin{figure}[h]
\centering
\includegraphics[width=0.70\textwidth]{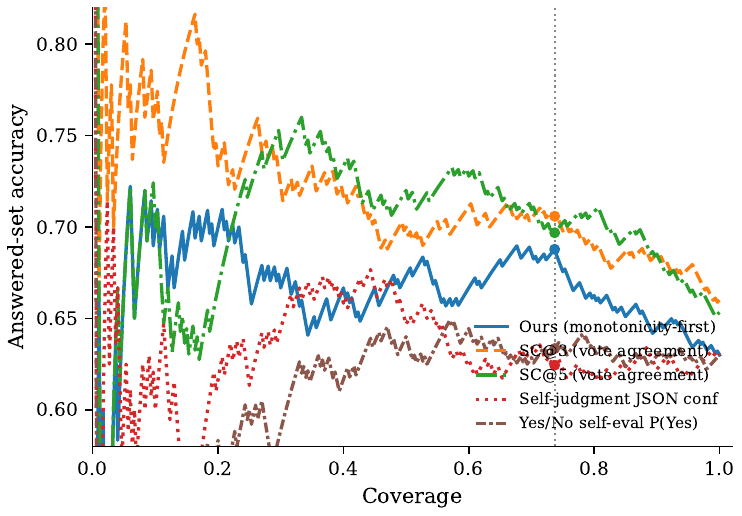}
\caption{Unified matched-coverage comparison on GSM8K ($n{=}300$):
our monotonicity-first ranking, SC@3/SC@5 vote-agreement rankings,
self-judgment confidence, and strict Yes/No self-evaluation confidence.
All methods are evaluated as answered-set accuracy versus coverage.}
\label{fig:matched_coverage_all}
\end{figure}

\subsection{Single-Trajectory Self-Judgment Baseline}
\label{app:self_judgment}

To address the reviewer request for a strong single-trajectory comparator, we
run a self-judgment baseline on the same 300 GSM8K problems using
Qwen2.5-7B-Instruct. For each problem, the model receives its own final answer
and emits one short JSON confidence score in $[0,100]$ via a verifier prompt
(one generation call per problem). We then rank by this confidence and report
the same selective metrics at target coverage $0.737$. Results are from
\texttt{figures/self\_judgment\_baseline.json}.

\paragraph{Protocol details (reproducible).}
Implementation is in \texttt{figures/run\_self\_judgment\_baseline.py}.
For each item, we construct a two-message chat prompt:
\texttt{system}: ``You are a strict math-answer verifier... output exactly one
line JSON \{\"confidence\": <0--100>\}.''
\texttt{user}: ``Question: ... Proposed final answer: ... Return only JSON.''
Decoding uses one verifier generation with \texttt{max\_new\_tokens=64},
temperature $0.0$ (greedy), and the model's native chat template.
Confidence extraction follows a deterministic rule: parse
\texttt{"confidence": number} if present; else parse the first numeric token
in output; else map lexical cues (``yes/correct'' $\rightarrow 0.75$,
``no/incorrect'' $\rightarrow 0.25$), with a final fallback $0.5$.

\paragraph{Evaluation alignment.}
To match our selective setup, self-judgment produces one score per problem,
then ranks all 300 items by score and reports Acc@coverage at
$k{=}\mathrm{round}(0.737\times 300)=221$ retained items. We also report
Spearman $\rho$ (score vs.\\ final correctness) and AURC under the same
risk-coverage definition used in the main text (lower is better).

Self-judgment yields Spearman $\rho{=}+0.019$, Acc@$73.7\%{=}62.4\%$, and
AURC $=0.368$, with mean verifier cost 7.6 tokens/problem (total
$242.3$ tokens/problem including the base chain). Despite low token overhead,
the ranking quality is weak and does not improve over random triage at matched
coverage, remaining substantially below trajectory-shape signals (AURC $0.311$;
Acc@$73.7\%$ $68.8\%$).

\paragraph{Strict Yes/No self-evaluation baseline.}
We also run a stricter single-token protocol in
\texttt{figures/run\_yesno\_self\_eval\_baseline.py}: the model must answer
``Yes'' or ``No'' to whether its own final answer is correct.
Confidence is taken as $P(\text{Yes})$ from first-token logits when available,
with deterministic fallback to parsed Yes/No output.
At 73.7\% coverage, this baseline reaches $63.3\%$ answered-set accuracy,
Spearman $\rho{=}{-}0.019$, and AURC $=0.395$ with only 2.0 verifier
tokens/problem on average (total $236.7$ including the base chain), indicating
very low cost but weak discrimination.

\subsection{Token Accounting Details and Fairness}
\label{app:token_accounting}

This section reconciles the two budget comparisons used in the paper:
(1) \emph{high-budget} reference points (SC@10/SC@40), and
(2) \emph{near-equal-budget} empirical baselines (SC@3/SC@5/ESC).

For our method, token usage decomposes as
\begin{equation}
T_{\text{ours}} = T_{\text{base-chain}} + m\,\bar{N}\,\bar{L}_{\text{short}},
\end{equation}
where $m{=}5$ is samples per prefix, $\bar{N}$ is mean number of prefixes,
and $\bar{L}_{\text{short}}$ is mean short-completion length.
On GSM8K (Qwen run, $n{=}300$), $\bar{N}{=}4.91$ and the measured base-chain
length is $\bar{T}_{\text{base-chain}}{=}234.7$ tokens/problem.

\begin{table}[h]
\centering
\caption{Unified token accounting on GSM8K ($n{=}300$).
``Measured'' values come from logged generations; ``configured'' values are
fixed decoding settings.}
\label{tab:token_accounting}
\small
\begin{tabular}{lcc}
\toprule
Quantity & Value & Type \\
\midrule
Base-chain tokens $\bar{T}_{\text{base-chain}}$ & $234.7$ & measured \\
Mean step prefixes $\bar{N}$ & $4.91$ & measured \\
Per-prefix samples $m$ & $5$ & configured \\
Short completion cap & $150$ & configured \\
Reported total $T_{\text{ours}}$ & $\approx1500$ & reported (main text) \\
\midrule
SC@3 tokens/problem & $831.4$ & measured \\
SC@5 tokens/problem & $1385.7$ & measured \\
ESC-sim tokens/problem & $673.6$ & measured \\
Self-judgment tokens/problem & $242.3$ & measured \\
\bottomrule
\end{tabular}
\end{table}

Interpreting these together: the ``$\approx1{,}500$'' figure is the operating
budget for our selective policy, while SC/ESC costs are directly measured at
full coverage. Thus, claims against SC@10/SC@40 are high-budget references,
and claims against SC@5/ESC are the strict empirical near-equal-budget
comparisons.

\end{document}